\documentclass[sigconf]{acmart}
\usepackage[linesnumbered,ruled,vlined]{algorithm2e}
\usepackage{bm}
\usepackage{tikz}
\usetikzlibrary{shapes}
\usetikzlibrary{positioning}
\usetikzlibrary{plotmarks}
\usetikzlibrary{patterns}
\usetikzlibrary{fit,calc}
\usetikzlibrary{arrows.meta}
\usepackage{microtype}
\usepackage{graphicx}
\usepackage{multirow}
\usepackage{enumitem}
\usepackage{booktabs}
\usepackage{adjustbox}
\usepackage{xspace}
\usepackage{multirow}
\usepackage{tabularx}
\usepackage{tikz}
\usepackage{subcaption}
\usetikzlibrary{positioning}
\usepackage{thmtools, thm-restate}
\usepackage{colortbl}
\usepackage{threeparttable}
\usepackage{float}
\newcommand{\wrt}{\text{w.r.t.\ }}

\AtBeginDocument{%
  }

\copyrightyear{2025}
\acmYear{2025}
\setcopyright{acmlicensed}\acmConference[MM '25]{Proceedings of the 33rd ACM International Conference on Multimedia}{October 27--31, 2025}{Dublin, Ireland}
\acmBooktitle{Proceedings of the 33rd ACM International Conference on Multimedia (MM '25), October 27--31, 2025, Dublin, Ireland}
\acmDOI{10.1145/3746027.3755302}
\acmISBN{979-8-4007-2035-2/2025/10}

\begin{document}

\title{Forward-Only Continual Learning}

\author{Jiao Chen}
\affiliation{%
  \institution{Shien-Ming Wu School of\\ Intelligent Engineering, South China University of Technology}
  \city{Guangzhou, Guangdong}
  \country{China}}
\email{202110190459@mail.scut.edu.cn}

\author{Jiayi He}
\affiliation{%
  \institution{Shien-Ming Wu School of\\ Intelligent Engineering, South China University of Technology}
  \city{Guangzhou, Guangdong}
  \country{China}}
\email{202164020171@mail.scut.edu.cn}

\author{Fangfang Chen}
\affiliation{%
  \institution{Shien-Ming Wu School of\\ Intelligent Engineering, South China University of Technology}
  \city{Guangzhou, Guangdong}
  \country{China}}
\email{wifannychen59@mail.scut.edu.cn}

\author{Zuohong Lv}
\affiliation{%
  \institution{Shien-Ming Wu School of\\ Intelligent Engineering, South China University of Technology}
  \city{Guangzhou, Guangdong}
  \country{China}}
\email{202220159664@mail.scut.edu.cn}

\author{Jianhua Tang}
\authornote{Corresponding author}
\affiliation{%
  \institution{Shien-Ming Wu School of\\ Intelligent Engineering, South China University of Technology}
  \city{Guangzhou, Guangdong}
  \country{China}}
\email{jtang4@e.ntu.edu.sg}

\renewcommand{\shortauthors}{Jiao Chen, Jiayi He, Fangfang Chen, Zuohong Lv, and Jianhua Tang}

\begin{abstract}
Catastrophic forgetting remains a central challenge in continual learning (CL) with pre-trained models. 
While existing approaches typically freeze the backbone and fine-tune a small number of parameters to mitigate forgetting, they still rely on iterative error backpropagation and gradient-based optimization, which can be computationally intensive and less suitable for resource-constrained environments.
To address this, we propose \textbf{FoRo}, a \textit{forward-only}, gradient-free continual learning method. 
FoRo consists of a lightweight prompt tuning strategy and a novel knowledge encoding mechanism, both designed without modifying the pre-trained model. Specifically, prompt embeddings are inserted at the input layer and optimized using the Covariance Matrix Adaptation Evolution Strategy (CMA-ES), which mitigates distribution shifts and extracts high-quality task representations. 
Subsequently, task-specific knowledge is encoded into a knowledge encoding matrix via nonlinear random projection and recursive least squares, enabling incremental updates to the classifier without revisiting prior data.
Experiments show that FoRo significantly reduces average forgetting and improves accuracy. Thanks to forward-only learning, FoRo reduces memory usage and run time while maintaining high knowledge retention across long task sequences. 
These results suggest that FoRo could serve as a promising direction for exploring continual learning with pre-trained models, especially in real-world multimedia applications where both efficiency and effectiveness are critical.
\end{abstract}

\ccsdesc[500]{Computing methodologies~Artificial intelligence~Continual learning}

\keywords{Continual Learning, Catastrophic Forgetting, Pre-trained Models, Knowledge Encoding, Prompt Learning}


\maketitle

\section{Introduction}\label{sec:introduction}
Humans possess the remarkable ability to incrementally acquire knowledge, adapt to changing environments, and leverage past experiences to solve new tasks ~\cite{leestella,cossu2024continual,li2024iob,10899876}. 
Continual learning (CL) aims to endow deep learning models with similar capabilities, enabling them to learn from non-stationary data streams without forgetting previous knowledge \cite{lomonaco2021avalanche,sun2023pilot,zhao2024continual,zhuang2022acil,wang2024comprehensive}. 

In real-world applications, CL systems are often deployed in resource-constrained environments, which impose strict limitations on computational cost, memory, model capacity, and data accessibility~\cite{our_comst,10102331,10899876,MMAL2024MM}. 
This makes it essential for a model to learn efficiently, preserve prior knowledge, and generalize well—all within a unified architecture.

Traditionally, CL methods train models from scratch, typically adopting residual networks (e.g., ResNet-based architectures) as backbones~\cite{zhuang2024ds,liu2024task,zhuang2022acil}.  
Representative approaches span three main categories: regularization-based methods~\cite{kirkpatrick2017overcoming,li2017learning}, replay-based methods~\cite{rebuffi2017icarl,yan2021dynamically,he2024continual}, and parameter isolation methods~\cite{mallya2018packnet}.

With the rapid advancement of Transformer-based architectures such as Vision Transformers (ViT)~\cite{dosovitskiy2020image}, continual learning (CL) has gradually shifted from training models from scratch to leveraging the representational power of pre-trained models (PTMs)~\cite{qu2025introducing,marczak2025magmax,sun2024mos,cao2024generative}.  
A common strategy is to freeze the pre-trained backbone and insert lightweight trainable modules, such as prompts \cite{jiang2020can}, which helps preserve the generalization ability of the model while mitigating forgetting.  
Another line of work applies small-step adaptation, either via low learning rates 
\cite{zhang2023slca} or limited iterations \cite{zhou2024continual,mcdonnell2024ranpac}, to slightly adjust the model for each task. Based on the adapted features, class prototypes are computed and directly used to replace the classifier weights, often in conjunction with a cosine classifier.

These existing methods mitigate forgetting by freezing the pre-trained backbone and fine-tuning a small number of task-specific parameters, they still rely on iterative error backpropagation and gradient-based optimization. This results in non-trivial computational overhead, limiting their applicability in resource-constrained environments such as edge devices~\cite{10878364} or real-time systems~\cite{10638762}. Moreover, despite their design to preserve prior knowledge, these gradient-based updates can still interfere with previously learned representations, leading to catastrophic forgetting~\cite{mcdonnell2024ranpac,zhuang2022acil}. This issue becomes particularly severe when tasks exhibit moderate similarity~\cite{yildiz2024investigating,he-etal-2025-self}.

Given this, we pose a central question:  
\textbf{Can continual learning be achieved without any reliance on backpropagation or gradient-based updates?}  
This reframes the problem from managing forgetting during parameter updates to designing a learning paradigm that avoids such updates altogether. By pursuing a purely forward optimization strategy, we aim to minimize computational cost while preventing destructive interference by design.

To this end, we propose \textbf{FoRo}, a forward-only, gradient-free continual learning method that leaves the pre-trained model entirely untouched. FoRo comprises a prompt-tuning strategy and a knowledge encoding mechanism. Specifically, prompt embeddings are appended to the input layer of a pre-trained model and optimized using the Covariance Matrix Adaptation Evolution Strategy (CMA-ES). This optimization helps bridge distributional gaps between tasks and accurately extract task-specific representations.
To retain past knowledge, FoRo encodes task-specific information into a knowledge encoding matrix via nonlinear random projection (NRP) and linear recursive least squares, enabling incremental classifier updates. The main novelty and contributions of our work are as follows:

1) We present FoRo, a forward-only, gradient-free continual learning method that operates without modifying the pre-trained model, making it well-suited for resource-constrained settings.

2) FoRo introduces two key components: prompt tuning via evolutionary optimization for task adaptation, and a knowledge encoding mechanism that incrementally updates the classifier without revisiting past data.

3) Experiments show that FoRo improves accuracy and reduces forgetting, while achieving lower memory and runtime costs through forward-only learning.

\section{Related Works}\label{sec:related works}
\subsection{PTMs-based Continual Learning}
Recent advances in CL with PTMs can be broadly categorized into two directions:

\textbf{Prompt-based methods} adapt PTMs to new tasks via lightweight trainable modules (i.e., prompts~\cite{jiang2020can}), while keeping the backbone frozen~\cite{kurniawan2024evolving,qin2024pearl}. A typical approach prepends learnable prompts to the input and optimizes them via cross-entropy loss to encode task-specific information~\cite{jia2022visual}.  
To improve adaptability, several works construct a prompt pool, which stores prompts as external memory and enables retrieval during both training and inference.  
DualPrompt~\cite{wang2022dualprompt} separates prompts into general and expert types for targeted retrieval.  
L2P~\cite{wang2022learning} introduces a query-key mechanism, assigning each prompt a learnable key and selecting prompts based on similarity with input features. 
MISA~\cite{kang2025advancing} further enhances prompt-based methods by leveraging pretraining data to initialize prompt parameters, thereby improving generalizability.  
Despite their efficiency, prompt-based methods face challenges such as suboptimal retrieval, forgetting due to shared parameters, and limited expressiveness from fixed-size prompt pools.

\textbf{Representation-based methods} leverage PTMs-extracted features to build task-specific classifiers. Some approaches~\cite{zhang2023slca,zhaosafe} apply dual learning rates—small for embedding layers and large for classifiers—combined with feature replay to recalibrate decision boundaries.  
ADAM~\cite{zhou2024continual} uses parameter-efficient modules (e.g., prompts or adapters) to fine-tune the PTMs and concatenates features from both the frozen and fine-tuned models to compute prototypes.  
RanPAC~\cite{mcdonnell2024ranpac} extends ADAM by applying random projection and online LDA to reduce inter-class correlation.  
LayUP~\cite{ahrens2023read} aggregates CLS tokens from the final transformer layers and trains an online LDA classifier on the concatenated features.  
While effective, these methods may introduce redundant representations, and the fusion of heterogeneous features can limit generalization. Early adaptation stages may also fail to fully bridge domain gaps in diverse downstream tasks.
\vspace{-0.1in}

\subsection{Forward-only learning}
Forward-only learning has received increasing attention in the communities of CL and PTMs. 
Its core idea is to update model parameters solely through forward propagation. Existing forward-only approaches can be categorized into three types: (1) perturbated model methods, which guide optimization by adding perturbations to model parameters and evaluating the resulting changes in output. This category includes forward gradient \cite{baydin2022gradients}, zeroth-order optimization \cite{malladi2023fine}, and evolutionary strategies \cite{vicol2023low}. Among them, CMA-ES \cite{hansen2016cma} is a representative evolutionary method that performs adaptive sampling and covariance adaptation in high-dimensional, non-convex spaces, and has been widely used in black-box optimization tasks. (2) Perturbated input methods, such as the Forward-Forward algorithm \cite{hinton2022forward}, adjust learning objectives based on activation discrepancies caused by input perturbations. (3) No perturbation methods \cite{ma2020hsic}, which compute parameter updates directly from hidden states or closed-form solutions, without any explicit perturbations.

Forward-only learning has been successfully applied to various PTMs-related tasks. In large language model optimization~\cite{sun2022black}, it offers an alternative to backpropagation. In test-time~\cite{niu2022efficient} and domain adaptation~\cite{bohdal2024feed}, forward-based updates enhance robustness under task switches and distribution shifts. \looseness=-1

\begin{figure*}[!t]
    \centering
    \includegraphics[width=0.65\linewidth]{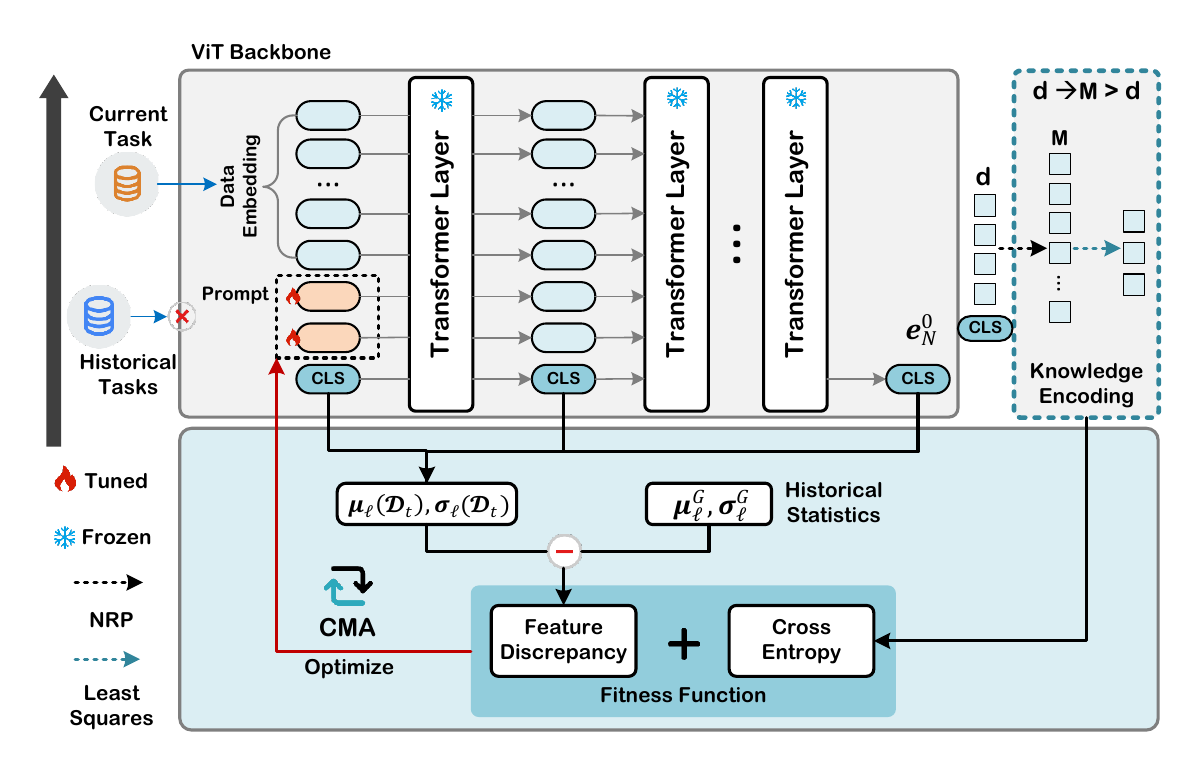}
    \caption{
    FoRo employs a prompt optimization strategy (i.e., CMA-ES) and a knowledge encoding mechanism to achieve continual learning across a sequence of tasks, leveraging a pre-trained ViT as the backbone. 
    }
    \label{fig:FoRo.}
    \vspace{-0.1in}
\end{figure*}
\section{Preliminary}\label{sec:prerequisites}

\subsection{Continual Learning Protocol}
CL is a strategy for training machine learning models to adapt to a sequence of changing tasks, each potentially introducing new data distributions. A series of tasks $\{\mathcal{D}_t\}_{t=1}^T$ is defined, where each task $\mathcal{D}_t$ consists of a set of data samples and corresponding labels, i.e., $\mathcal{D}_t = \{(\mathbf{x}_i^t, y_i^t)\}_{i=1}^{|\mathcal{D}_t|}$, where $\mathbf{x}_i^t$ is the input data and $y_i^t$ is the corresponding label. The goal in each task stage $t$ is for the model to learn the new task's data while addressing catastrophic forgetting on previous tasks. \looseness=-1

Class-Incremental Learning (CIL) is an important branch of CL. 
In CIL, each task $\mathcal{D}_t$ introduces an independent set of categories, i.e., $y_i^t \in \mathcal{Y}_t$ and $\mathcal{Y}_t \cap \mathcal{Y}_j = \emptyset$ for all $t \neq j$. 
According to the goals of CL, the objective for CIL is to fit a model $f(\cdot)$, and minimize the cumulative expected risk across the entire sequence of tasks:
\begin{equation}
    f^* = \arg \min_{f\in\mathcal{H}} \sum_{t=1}^T \mathbb{E}_{(\mathbf{x}_i^t,y_i^t) \sim \mathcal{D}_t} \left[ \mathbb{I}[f(\mathbf{x}_i^t) \neq y_i^t] \right],
\end{equation}
where $\mathcal{H}$ is the hypothesis space, $\mathbb{I}(\cdot)$ is the indicator function which outputs 1 if the expression holds and 0 otherwise.

\subsection{Vision Transformer}
ViT is a specialized neural network designed for image processing. Unlike conventional CNNs that apply filters directly to image regions, ViT divides an image into patches, treating each patch as a token akin to word processing in NLP. This enables ViT to utilize the powerful self-attention mechanism, which captures complex relationships across image segments effectively. \looseness=-1

As depicted in Fig.~\ref{fig:FoRo.}, a ViT processes images through multiple layers. Each layer transforms a set of input patch embeddings, supplemented with positional information, into refined embeddings. The classification output is typically derived from the embedding of a special token (denoted as [CLS]) added to the sequence of patch embeddings.
Mathematically, consider a ViT with \( N \) layers. Each layer \( L_\ell \) transforms the input patch embeddings \(\mathbf{E}_{\ell-1}\) from the previous layer. Let \(\mathbf{E}_\ell = \{\bm{e}_\ell^j\}_{j=0}^{m}\) represent these patch embeddings, where \( m \) is the number of image patches and \( \bm{e}_\ell^0 \) denotes an extra learnable classification token ([CLS]) for the \( \ell \)-th layer.
The transformation process through the layers of ViT can be summarized by the following equations:
\begin{align}
    \textbf{\text{E}}_\ell &= L_\ell(\textbf{\text{E}}_{\ell-1}), \quad \ell = 1, \ldots, N, \\
    \hat{y} &= \mathbf{W}^{\top}\bm{e}^0_N,
\end{align}
where $\mathbf{W}\in\mathbb{R}^{d\times |\cup_{t}^{T}\mathcal{Y}_t|}$ is the classification head and $d$ denotes the embedding dim.


\section{Forward-Only Continual Learning}
FoRo comprises two key components: a lightweight prompt tuning strategy and a knowledge encoding mechanism, both designed without modifying the parameters of the pre-trained model. 

\subsection{Prompt Tuning with CMA-ES}

To enable continual learning without altering the pre-trained backbone, FoRo prepends a set of learnable prompts \( \textbf{p} \in \mathbb{R}^{P \times d} \) to the input patch embeddings \( \mathbf{x}_i^t \), where \( P \) is the number of prompts and \( d \) is the embedding dimension. The concatenated sequence \( [\textbf{p}, \mathbf{x}_i^t] \) is fed into the frozen ViT model. These prompts are task-adaptive parameters optimized for each task \( \mathcal{D}_t \) to extract relevant features while preserving previously acquired knowledge.

Existing methods such as L2P~\cite{wang2022learning} and RanPAC~\cite{mcdonnell2024ranpac} introduce learnable prompts for adapting PTMs to sequential tasks. However, they typically rely on iterative, gradient-based updates, which can face two primary limitations in CL settings:

\begin{itemize}
    \item \textit{Risk of local optima}:
    Since these methods often optimize prompts with incomplete historical data, they may converge to task-specific local minima rather than a more global solution that balances all tasks~\cite{zhuang2022acil}.
    \item \textit{High computational cost}:
    Repeated backpropagation in high-dimensional prompt spaces demands significant memory and compute resources, making these methods less efficient, particularly when many tasks arrive sequentially~\cite{niu2024test}.
\end{itemize}

To alleviate these challenges, FoRo adopts an \textit{evolution strategy}, specifically the Covariance Matrix Adaptation Evolution Strategy (CMA-ES)~\cite{hansen2016cma}, a gradient-free, black-box optimization algorithm well-suited for high-dimensional, non-convex search spaces. 
CMA-ES maintains a multivariate Gaussian distribution over candidate prompts and iteratively refines this distribution by sampling, evaluating, and updating, enabling efficient global exploration without requiring gradients or differentiable objectives.

At each iteration during the continual learning process for task \( \mathcal{D}_t \), a population of candidate prompts is sampled as:
\begin{equation}\label{cma_evolution}
    \mathbf{p}_k^{(t)} \sim \mathbf{m}^{(t)} + \tau^{(t)} \mathcal{N}(\mathbf{0}, \mathbf{\Sigma}^{(t)}),
\end{equation}
where \( k \) indexes the population, \( \mathbf{m}^{(t)} \in \mathbb{R}^{P \times d} \) is the mean prompt vector, \( \tau^{(t)} \in \mathbb{R} \) is the global step size, and \( \mathbf{\Sigma}^{(t)} \in \mathbb{R}^{(Pd) \times (Pd)} \) is the covariance matrix. This formulation allows CMA-ES to learn both the direction and scale of updates in the high-dimensional prompt space.

\textbf{Fitness Function.}
CMA-ES requires a metric to assess the quality of each candidate prompt. Specifically, at each iteration, it samples a set of candidate prompts from its current distribution, evaluates them against a \emph{fitness function}, and then updates the distribution parameters (mean and covariance) based on the evaluation results. By continually cycling through sampling, evaluating, and updating, CMA-ES incrementally refines its search toward more effective prompt solutions.

As illustrated in Figure~\ref{fig:FoRo.}, we define a task-specific fitness function \( \mathcal{L}(\mathbf{p}; \mathcal{D}_t) \), which guides the evolution process. A naive option would be the cross-entropy loss. However, due to task distribution shifts in CL, we propose a regularized fitness function incorporating multi-layer \textit{activation discrepancy}~\cite{chencross}, which encourages consistency of representations across tasks:
\vspace{-0.1in}

\small
\begin{equation}\label{fitness function}
\begin{split}
    \mathcal{L}(\mathbf{p}; \mathcal{D}_t) = & \underbrace{\sum_{(\mathbf{x}_i^t, y_i^t) \in \mathcal{D}_t} \sum_{y_i^t \in \mathcal{Y}_t} -y_i^t \log \hat{y}_i^t}_{(\textbf{a})} \\
    & + \underbrace{\lambda \sum_{\ell=1}^{N} \left( \|\bm{\mu}_\ell(\mathcal{D}_t) - \bm{\mu}_\ell^G(t-1)\|_2 + \|\bm{\sigma}_\ell(\mathcal{D}_t) - \bm{\sigma}_\ell^G(t-1)\|_2 \right)}_{(\textbf{b})},
\end{split}
\end{equation}
\normalsize
where term (\textbf{a}) is the cross-entropy loss, ensuring the model captures task-specific knowledge.  
Term (\textbf{b}) is the activation discrepancy regularizer, which aligns the statistical distribution (mean and standard deviation) of the current task’s intermediate representations with those of previous tasks, promoting generalization across the task sequence. Specifically, \( \bm{\mu}_\ell(\mathcal{D}_t) \) and \( \bm{\sigma}_\ell(\mathcal{D}_t) \) denote the mean and standard deviation of layer-\( \ell \) CLS activations for the current task, while \( \bm{\mu}_\ell^G(t-1) \) and \( \bm{\sigma}_\ell^G(t-1) \) store historical statistics.

The historical mean \( \bm{\mu}_\ell^G(t-1) \) is updated using an interpolation method to smoothly integrate new task statistics. We define the update of \( \bm{\mu}_\ell^G \) in iteration \( t \) as:
\begin{equation}
    \bm{\mu}_\ell^G(t) = \alpha \bm{\mu}_\ell(\mathcal{D}_t) + (1 - \alpha) \bm{\mu}_\ell^G(t - 1),
\end{equation}
where \( \bm{\mu}_\ell(\mathcal{D}_t) \) is the mean of the activations for the \( \ell \)-th layer calculated over the \( t \)-th task batch \( \mathcal{D}_t \), and \( \alpha \in [0, 1] \) is a moving average factor, set to 0.1 in our experiments. The same interpolation method is applied to update the standard deviation \( \bm{\sigma}_\ell^G(t-1) \).

Intuitively, this regularization serves as feature-level knowledge preservation: by aligning activation statistics across layers, it encourages the learned prompts to produce representations consistent with earlier tasks.  
Unlike parameter-based regularization (e.g., L2 or Fisher), activation discrepancy directly enforces output-level consistency, making it more robust in frozen-backbone settings. It can thus be seen as a lightweight alternative to rehearsal, preserving past task representations without storing data.

The fitness value $v_k$ of each candidate prompt is then evaluated using the defined fitness function (Eqn.~\ref{fitness function}). 
Based on the fitness scores, the distribution parameters \( \mathbf{m}^{(t)} \), \( \tau^{(t)} \), and \( \mathbf{\Sigma}^{(t)} \) are updated to maximize the likelihood of successful candidate prompts in future iterations. 
Based on the fitness value $\{v_k\}_{k=1}^K$, CMA-ES updates the distribution parameters as follows: the mean $\mathbf{m}^{(t)}$ shifts toward top candidates, particularly the one with the highest fitness value, denoted as $v_{\text{best}}$, which represents the optimal prompt.The step size $\mathbf{\tau}^{(t)}$ adapts based on dispersion for balance between exploration and exploitation, and the covariance $\mathbf{\Sigma}^{(t)}$ aligns with high-fitness regions to capture solution geometry.
This process iteratively improves the prompt parameters for effective adaptation to new tasks in a CL method.

\begin{figure*}[!t]
    \centering
    \includegraphics[width=0.65\linewidth]{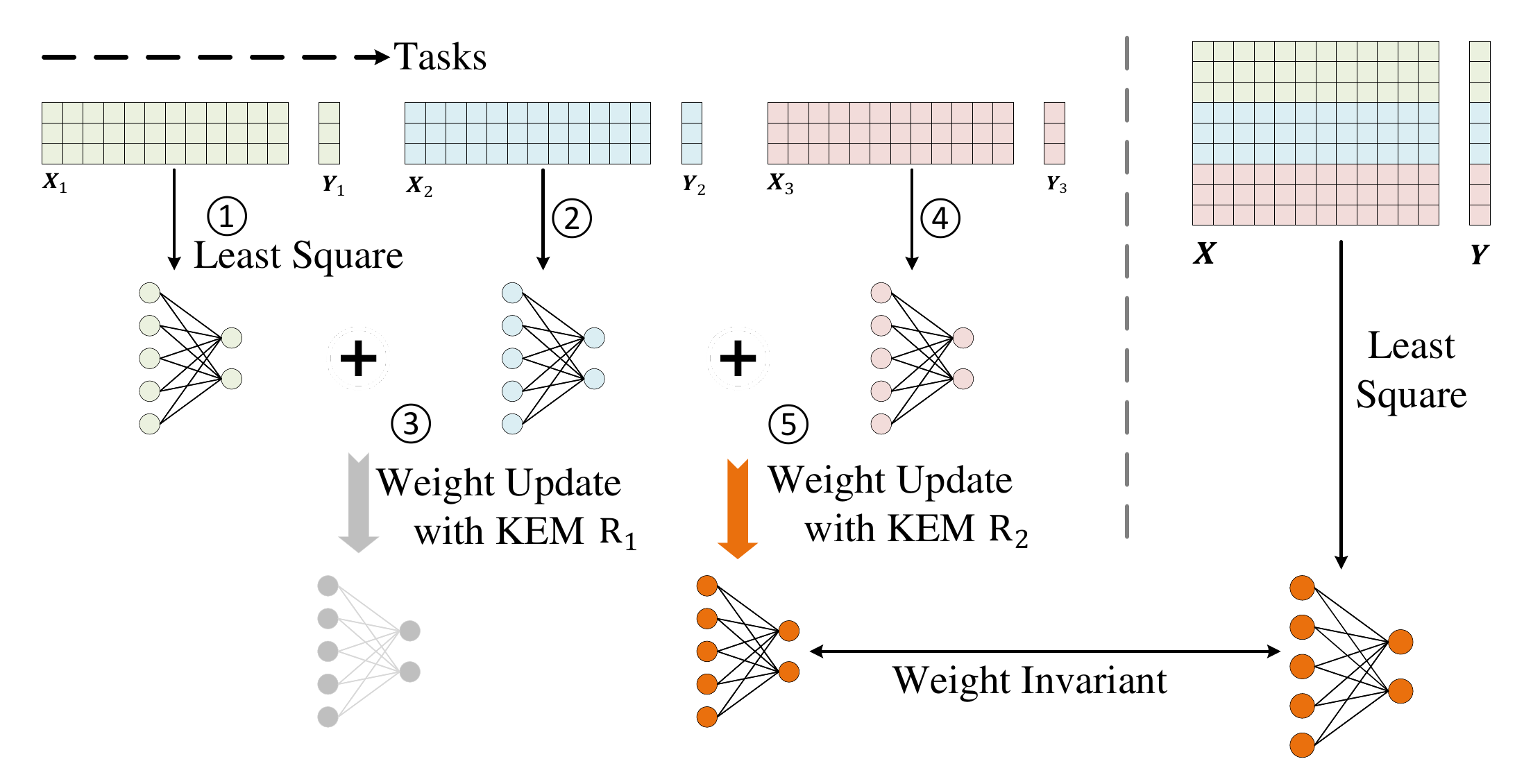}
    \caption{
    Knowledge Encoding.
    In CL, the current task knowledge is encoded into the KEM and then updates the weight of classfier. This approach yields weights that are equivalent to those obtained using all samples for weight computation.
    }
    \label{fig:kem}
    \vspace{-0.1in}
\end{figure*}

\subsection{Knowledge Encoding} \label{know. encoding}
While prompt tuning enables rapid task adaptation without altering the backbone network, effective long-term knowledge retention across sequential tasks requires both expressive task representations and continual updates of classifier weights.  
To address this, FoRo introduces a second key component: a \textit{Knowledge Encoding} mechanism, which, as illustrated in Figure~\ref{fig:kem}, incrementally updates the classifier's weights in a forward-only fashion without revisiting past data.
This forward update strategy is inspired in part by ACIL~\cite{zhuang2022acil,L3A2025ICML,GKEAL2023CVPR}, which proposes an analytic formulation to recursively adjust classifier weights while achieving strong memorization without replay.

\textbf{Knowledge Encoding Matrix (KEM).}
Consider the final classification layer as a linear mapping from the backbone's embeddings (e.g., CLS tokens) to class logits. Let 
\(\mathbf{X} \in \mathbb{R}^{n \times d}\) be feature embeddings (\(n\) samples, \(d\) feature dimension), and 
\(\mathbf{Y} \in \mathbb{R}^{n \times c}\) the corresponding labels (\(c\) classes). 
A regularized least squares formulation finds the optimal classifier weights 
\(\mathbf{W} \in \mathbb{R}^{d \times c}\):
\begin{equation}
    \label{eq:ridge}
    \min_{\mathbf{W}} \, \|\mathbf{Y} - \mathbf{X} \mathbf{W}\|_F^2 + \gamma \|\mathbf{W}\|_F^2,
\end{equation}
whose closed-form solution is
\begin{equation}
    \label{eq:cls-solution}
    \mathbf{\hat{W}} 
    = (\mathbf{X}^{\top}\mathbf{X} + \gamma \mathbf{I})^{-1} \mathbf{X}^{\top}\mathbf{Y},
\end{equation}
where \(\gamma\) is a regularization parameter and \(\mathbf{I} \in \mathbb{R}^{d \times d}\) is the identity matrix. 
Recomputing \((\mathbf{X}^{\top}\mathbf{X} + \gamma \mathbf{I})^{-1}\) from scratch each time a new task arrives is costly, and storing all past data \(\mathbf{X}, \mathbf{Y}\) is impractical for continual learning.

We therefore introduce a \textit{Knowledge Encoding Matrix (KEM)}:
\begin{equation}
    \mathbf{R} = (\mathbf{X}^{\top} \mathbf{X} + \gamma \mathbf{I})^{-1} \in \mathbb{R}^{d \times d},
\end{equation}
which serves as a compact representation of accumulated task information. 
Crucially, \(\mathbf{R}\) can be \emph{recursively updated} as new data arrive, without retaining all previous inputs, thus enabling forward-only classifier updates while preserving the optimality of the least squares solution.

\textbf{Efficient Recursive Update via Woodbury Identity.}  
The Woodbury identity yields an efficient rule for incrementally updating matrix inverses. 
Let \(\mathbf{A} \in \mathbb{R}^{d \times d}\) be an invertible matrix, and 
\(\mathbf{U} \in \mathbb{R}^{d \times n}, \mathbf{V} \in \mathbb{R}^{n \times d}, \mathbf{C} \in \mathbb{R}^{n \times n}\). Then:
\begin{equation}
    (\mathbf{A} + \mathbf{U}\mathbf{C}\mathbf{V})^{-1} 
    = 
    \mathbf{A}^{-1} 
    - 
    \mathbf{A}^{-1} \mathbf{U} 
    \bigl(\mathbf{C}^{-1} + \mathbf{V}\mathbf{A}^{-1}\mathbf{U}\bigr)^{-1} 
    \mathbf{V} \mathbf{A}^{-1}.
\end{equation}
When new task data \(\mathbf{X}_t \in \mathbb{R}^{n_t \times d}\) arrive, we update \(\mathbf{R}_{t-1}\) as:
\begin{equation} 
\label{eq:kem-update}
\mathbf{R}_t 
= 
\mathbf{R}_{t-1} 
- 
\mathbf{R}_{t-1} \mathbf{X}_t^{\top} 
\bigl(\mathbf{I}_{n_t} + \mathbf{X}_t \mathbf{R}_{t-1} \mathbf{X}_t^{\top}\bigr)^{-1} 
\mathbf{X}_t \mathbf{R}_{t-1},
\end{equation}
starting from \(\mathbf{R}_0 = (\gamma \mathbf{I})^{-1}\). 
Here, \(\mathbf{I}_{n_t}\) is the \(n_t \times n_t\) identity matrix. 
This can be seen by setting 
\(\mathbf{A} = \mathbf{R}_{t-1}^{-1}, \mathbf{U} = \mathbf{X}_t^{\top}, \mathbf{V} = \mathbf{X}_t, \mathbf{C} = \mathbf{I}_{n_t}\) in the Woodbury formula.

\textbf{Incremental Weight Updates.}  
Using the updated KEM \(\mathbf{R}_t\), we can efficiently update classifier weights without revisiting all historical data. 

\textbf{Step 1}: Note that the least squares solution after seeing all tasks up to \(t\) can be written as:
\begin{equation}
    \mathbf{W}_t 
= 
\bigl(\mathbf{X}_{1:t}^{\top}\mathbf{X}_{1:t} + \gamma \mathbf{I}\bigr)^{-1}
\mathbf{X}_{1:t}^{\top}\mathbf{Y}_{1:t}.
\end{equation}

\textbf{Step 2}: Since \(\mathbf{R}_t = (\mathbf{X}_{1:t}^{\top}\mathbf{X}_{1:t} + \gamma \mathbf{I})^{-1}\), we have:
\begin{equation}
    \mathbf{W}_t 
= 
\mathbf{R}_t \, \mathbf{X}_{1:t}^{\top} \, \mathbf{Y}_{1:t}.
\end{equation}

\textbf{Step 3}: Separating past and current data, 
\begin{equation}
    \mathbf{W}_t 
= 
\mathbf{R}_t
\bigl(
  \mathbf{X}_{1:t-1}^{\top}\mathbf{Y}_{1:t-1} 
  + 
  \mathbf{X}_t^{\top}\mathbf{Y}_t
\bigr).
\end{equation}

\textbf{Step 4}: Defining \(\mathbf{W}_{t-1} = \mathbf{R}_{t-1}\,\mathbf{X}_{1:t-1}^{\top}\,\mathbf{Y}_{1:t-1}\) yields:
\begin{equation}
    \mathbf{W}_t 
= 
\mathbf{W}_{t-1} 
+ 
\mathbf{R}_t 
\mathbf{X}_t^{\top}
\bigl(\mathbf{Y}_t - \mathbf{X}_t \mathbf{W}_{t-1}\bigr),
\end{equation}
leading to the final recursive rule:
\begin{equation}
\label{eq:incremental-weight}
\mathbf{\hat{W}}_t 
= 
\mathbf{\hat{W}}_{t-1}
+ 
\mathbf{R}_t \mathbf{X}_t^{\top}
\bigl(\mathbf{Y}_t - \mathbf{X}_t \mathbf{\hat{W}}_{t-1}\bigr).
\end{equation}

\textbf{Extending the Classifier for New Classes.}
In a class-incremental setting, if new tasks introduce \emph{novel classes} beyond the existing ones, the classifier weight matrix 
\(\mathbf{W}\in\mathbb{R}^{d \times c}\) must be extended along its columns to accommodate additional classes (i.e., from \(c\) to \(c + \Delta c\)). 
By contrast, the KEM 
\(\mathbf{R}\in\mathbb{R}^{d \times d}\) 
depends only on the feature dimension \(d\); if the backbone’s feature dimension remains the same, \(\mathbf{R}\) remains \(\mathbb{R}^{d \times d}\) and does not require resizing. 
Thus, newly added columns in \(\mathbf{W}\) will be learned incrementally through the same least squares update process, ensuring scalability in class-incremental learning.

\textbf{Nonlinear Random Projection.}
Although prompt tuning and classifier updating ensure task adaptation and knowledge retention, the quality of the input features remains crucial. To further enhance the separability of task-specific features—especially under a frozen backbone—we introduce a nonlinear random projection (NRP) module. This module improves the discriminative power of the CLS-token embeddings without adding trainable parameters, maintaining the forward-only design of FoRo.

Specifically, a nonlinear activation function $\phi(\cdot)$ is defined, and for a given feature sample $\bm{e}_N^0$, the transformed feature representation $\bm{h}$ of length $M$ is obtained by $\bm{h} := \phi(\bm{e}_N^0 \mathbf{W}_{rp})$, where $\mathbf{W}_{rp}$ denotes the random projection matrix. This transformation process is illustrated in Fig.~\ref{fig:FoRo.}. \looseness=-1

By applying a nonlinear activation function $\phi$ after random projection, enriched nonlinear feature interactions are created. 
These interactions enhance the class separability of the features, contributing to more robust task-specific representations. 
In FoRo, the new feature representation $\bm{h}$ replaces the original features $\bm{e}_N^0$ for subsequent task computations, thereby maintaining model performance and stability across sequential tasks. 

Details of the NRP are discussed in the Section~\ref{sec:nrp}.
The pseudocode of FoRo is summarized in Algorithm~\ref{alg}.

\begin{algorithm}[tb]
\caption{Forward-Only Continual Learning}
\label{alg}
\SetAlgoLined
\DontPrintSemicolon
\KwIn{Sequence of tasks $\{\mathcal{D}_1, \ldots, \mathcal{D}_T\}$, 
   historical statistics $\{\bm{\mu}_\ell^G, \bm{\sigma}_\ell^G\}_{\ell=0}^N$, population size $K$.}
\KwOut{Optimized prompts $\bm{\textbf{p}}$ and weight matrix $\hat{\mathbf{W}}$.}

Initialize $\mathbf{m}^{(0)} = \mathbf{0}$, $\mathbf{\Sigma}^{(0)} = \mathbf{I}$, $\tau^{(0)} = \mathbf{1}$ in Eqn.~(\ref{cma_evolution}).

\For{$t = 1$ \KwTo $T$}{
    Sample $K$ prompt solutions $\{\bm{\textbf{p}}_k^{t}\}_{k=1}^{K}$ by Eqn.~(\ref{cma_evolution}).
    \For{$k = 1$ \KwTo K}{
        Calculate all layer's \text{[CLS]} features $\{\bm{e}_\ell^0\}_{\ell=1}^N$ using Eqn. (2) with a batch of $\mathcal{D}_t$.
        Calculate fitness value $v_k$ per Eqn.~(\ref{fitness function}).
    }
    Update $\mathbf{m}^{(t)}$, $\mathbf{\Sigma}^{(t)}$, $\tau^{(t)}$ according to $\{v_k\}_{k=1}^K$ using the CMA-ES algorithm \cite{hansen2016cma}.\;
    Obtain the feature matrix of $\mathcal{D}_t$ by the best prompt $\bm{\textbf{p}}^{t}_{\text{best}}$ with the highest fitness value $v_{\text{best}}^t.$\;
    Update KEM $\mathbf{R}_t$ using Eqn.~(\ref{eq:kem-update}).\;
    Update weight matrix $\hat{\mathbf{W}}_{t}$ using Eqn.~(\ref{eq:incremental-weight}).\;
}
\end{algorithm}
\vspace{-0.1in}

\subsection{Complexity Analysis}\label{sec:complexity}

The space complexity of FoRo is primarily determined by two matrices: the covariance matrix used in CMA-ES, which has a size of \( D_1^2 \), where \( D_1 \) is the dimension of the token (typically 768 in ViT), and the knowledge encoding matrix $\mathbf{R}$, which has a size of \( D_2^2 \), where \( D_2 \) is obtained through NRP and is set to 8192. The rationale behind this choice is further discussed in Section~\ref{sec:nrp}. Therefore, the total space complexity of FoRo is \( \mathcal{O}(D_1^2 + D_2^2) \). Overall, the number of trainable parameters is relatively small, and gradient computation is not required, making FoRo highly efficient in memory usage.

For the time complexity, it is mainly dominated by the update operations of these two matrices. The complexity of updating the covariance matrix is approximately \( \mathcal{O}(D_1^3) \). 
The complexity of updating the matrix $\mathbf{R}$ is mainly determined by matrix multiplication, approximately $\mathcal{O}(D_2^2)$. For $\mathbf{W}$, it is determined by both matrix multiplication and matrix inversion, also approximately $\mathcal{O}(D_2^2)$, as the batch size is small and the class number grows incrementally but remains bounded (e.g., up to 100 for CIFAR-100). Thus, the overall computational complexity for knowledge encoding mechanism is $\mathcal{O}(D_2^2)$.
Thus, the total time complexity of FoRo is approximately: \( \mathcal{O}(D_1^3 + D_2^2) \).

\section{Experiments}
\subsection{Basic Setups}
Building on previous works~\cite{zhou2024continual,mcdonnell2024ranpac}, we employ the ViT-B/16 architecture \cite{dosovitskiy2020image}, which is pre-trained on ImageNet-1K and features 12 Self-Attention blocks with channel dimensions of 768. \looseness=-1

\textbf{Dataset and Models.}
We conduct experiments on three datasets: CIFAR-100, ImageNet-R (various artistic renditions of 200 ImageNet classes), and CUB-200 \cite{wah2011caltech}. The latter two are considered to have a significant domain gap from ImageNet \cite{zhou2024continual,mcdonnell2024ranpac,ahrens2023read}.

\textbf{Compared Methods.}
We compare FoRo with seven CL methods, covering regularization, replay, prompt tuning, and prototype-based strategies:
(1) \textbf{Fine-tune (FT)}: Continuously trains a pre-trained model on new tasks, updating all parameters.
(2) \textbf{LwF} \cite{li2017learning}: Maintains performance by aligning new and old task knowledge using distillation loss.
(3) \textbf{DSG} \cite{he2024continual}: Mitigates catastrophic forgetting through generative replay using diffusion models to generate old task data.
(4) \textbf{L2P} \cite{wang2022learning}: Introduces visual prompt tuning in CL, using a pre-trained ViT and a prompt pool to select instance-specific prompts.
(5) \textbf{LayUP} \cite{ahrens2023read} leverages rich, fine-grained information from multiple layers of PTMs to enhance prototype-based continual learning.
(6) \textbf{ADAM} \cite{zhou2024continual}: ADAM initially employs parameter-efficient fine-tuning (e.g., Adapter) of the pre-trained model using data from the first incremental task. Upon completion of the first incremental phase, the entire backbone network is frozen, and a prototype-based approach is adopted to construct the classifier.
(7) \textbf{RanPAC}~\cite{mcdonnell2024ranpac}: RanPAC enhances pre-trained model fine-tuning by adjusting class prototypes using the Gram matrix, thereby improving robustness against overlapping class distributions through second-order statistical measures. \looseness=-1

\textbf{Implementation Details.}
The model incrementally learns all categories in the dataset over a sequence of $T$ phases. In each phase, the categories are mutually exclusive, ensuring that no category overlaps across phases. By default, the number of categories in each task $\mathcal{D}_t$ is kept equal. Experiments are conducted using three seeds, and the results are reported as averages.
We set the dimension of the NRP to 8192 and the regularization factor \( \gamma \) to 0.1. This configuration was empirically selected to strike a balance between model performance and computational cost, as further discussed in Section~\ref{sec:nrp}. \looseness=-1

\textbf{Evaluation Metrics.}
We use the following metrics to evaluate model performance \cite{mirzadeh2020understanding}:
(1) \textbf{Average Accuracy}: The average test accuracy over all tasks when the number of tasks is \(T\). It is defined as: \(\overline{\bm{\text{A}}}_T = \frac{1}{T} \sum_{t=1}^{T} a_{T,t}\), where \(a_{T,t}\) denotes the test accuracy of task \(t\) after learning \(T\) tasks. 
(2) \textbf{Average Forgetting}: The average forgetting over all tasks when the number of tasks is \(T\). Forgetting is defined as the decrease in performance at each task between their peak accuracy and their accuracy after the CL experience has finished. It is defined as:
\begin{equation}
    \overline{\bm{\text{F}}} = \frac{1}{T-1} \sum_{t=1}^{T-1} \max_{j \in \{1, \ldots, T-1\}} (a_{j,t} - a_{T,t}),
\end{equation}
where \(a_{j,t}\) is the test accuracy of task \(t\) after learning \(j\) tasks, and \(a_{T,t}\) is the test accuracy of task \(t\) after learning all \(T\) tasks. \looseness=-1

\begin{table*}[ht]
\centering
\caption{The performance comparison of different methods. Mean ± Standard Deviation over 3 runs.}
\label{tab:Performance_Comparision}
\resizebox{1.0\linewidth}{!}{
\begin{tabular}{@{}lcccccccccccccc@{}}
\toprule
\multirow{3}{*}{Method} & \multicolumn{4}{c}{CIFAR-100} & \multicolumn{4}{c}{ImageNet-R} & \multicolumn{4}{c}{CUB-200} & \multicolumn{2}{c}{\multirow{2}{*}{Average}} \\ \cmidrule(lr){2-13}
& \multicolumn{2}{c}{T = 5} & \multicolumn{2}{c}{T = 25} & \multicolumn{2}{c}{T = 5} & \multicolumn{2}{c}{T = 25} & \multicolumn{2}{c}{T = 5} & \multicolumn{2}{c}{T = 25} & \multicolumn{2}{c}{} \\ \cmidrule(l){2-15} 
& $\overline{\bm{\text{A}}}_5$ (\%, $\uparrow$) & $\overline{\bm{\text{F}}}$ (\%, $\downarrow$) & $\overline{\bm{\text{A}}}_{25}$ & $\overline{\bm{\text{F}}}$ & $\overline{\bm{\text{A}}}_5$ & $\overline{\bm{\text{F}}}$ & $\overline{\bm{\text{A}}}_{25}$ & $\overline{\bm{\text{F}}}$ & $\overline{\bm{\text{A}}}_5$ & $\overline{\bm{\text{F}}}$ & $\overline{\bm{\text{A}}}_{25}$ & $\overline{\bm{\text{F}}}$ & $\overline{\bm{\text{A}}}$ & $\overline{\bm{\text{F}}}$ \\ \midrule
FT & 73.2 \tiny($\pm$ 3.09) & 4.5 \tiny($\pm$ 0.86) & 55.7 \tiny($\pm$ 1.81) & 32.7 \tiny($\pm$ 2.41) & 61.4 \tiny($\pm$ 0.96) & 5.8 \tiny($\pm$ 0.51) & 39.1 \tiny($\pm$ 0.65) & 36.4 \tiny($\pm$ 0.43) & 60.1 \tiny($\pm$ 1.28)  & 16.3 \tiny($\pm$ 0.91) & 54.3 \tiny($\pm$ 1.32) & 25.6 \tiny($\pm$ 0.54) & 57.3 & 20.2 \\
LwF & 83.3 \tiny($\pm$ 1.46) & 9.1 \tiny($\pm$ 0.72) & 73.6 \tiny($\pm$ 1.36) & 10.7 \tiny($\pm$ 2.50) & 74.4 \tiny($\pm$ 0.86) & 7.7 \tiny($\pm$ 0.53) & 58.0 \tiny($\pm$ 0.78) & 13.2 \tiny($\pm$ 0.39) & 67.3 \tiny($\pm$ 0.62) & 10.1 \tiny($\pm$ 0.30) & 60.9 \tiny($\pm$ 0.75) & 13.7 \tiny($\pm$ 0.28) & 69.6 & 10.8 \\
DSG & 86.8 \tiny($\pm$ 1.82) & 5.8 \tiny($\pm$ 0.79) & 84.1 \tiny($\pm$ 0.86) & 6.1 \tiny($\pm$ 1.95) & 71.1 \tiny($\pm$ 1.06) & \underline{3.6 \tiny($\pm$ 0.58)} & 66.3 \tiny($\pm$ 0.74) & 8.7 \tiny($\pm$ 0.41) & 75.2 \tiny($\pm$ 0.85) & 6.9 \tiny($\pm$ 0.70) & 69.8 \tiny($\pm$ 0.73) & 5.8 \tiny($\pm$ 0.35) & 75.6 & 6.2 \\
L2P & 85.8 \tiny($\pm$ 1.58) & 7.7 \tiny($\pm$ 1.76) & 81.3 \tiny($\pm$ 1.69) & 10.2 \tiny($\pm$ 0.72) & 72.9 \tiny($\pm$ 1.37) & 5.8 \tiny($\pm$ 0.33) & 68.0 \tiny($\pm$ 0.97) & 7.6 \tiny($\pm$ 0.42) & 82.1 \tiny($\pm$ 1.05) & 5.7 \tiny($\pm$ 0.83) & 75.4 \tiny($\pm$ 1.14) & 8.7 \tiny($\pm$ 0.57) & 77.6 & 7.6 \\
LayUP & \underline{90.2 \tiny($\pm$ 0.47)} & 3.8 \tiny($\pm$ 0.49) & \underline{84.9 \tiny($\pm$ 0.92)} & 6.9 \tiny($\pm$ 0.66) & 78.0 \tiny($\pm$ 0.94) & 3.9 \tiny($\pm$ 1.03) & 74.1 \tiny($\pm$ 0.75) & \underline{6.3 \tiny($\pm$ 0.50)} & \textbf{90.1 \tiny($\pm$ 0.63)} & 4.1 \tiny($\pm$ 0.31)  & 83.3 \tiny($\pm$ 0.42)  & 5.9 \tiny($\pm$ 0.17)  & 83.4 & 5.2 \\
ADAM & \textbf{90.9 \tiny($\pm$ 1.32)} & \underline{3.6 \tiny($\pm$ 1.08)} & 84.7 \tiny($\pm$ 0.79) & 7.1 \tiny($\pm$ 0.67) & \textbf{79.4 \tiny($\pm$ 0.63)} & 3.8 \tiny($\pm$ 0.35) & 73.8 \tiny($\pm$ 0.94) & 6.8 \tiny($\pm$ 0.22) & 89.1 \tiny($\pm$ 0.71) & \underline{3.7 \tiny($\pm$ 0.35)} & \underline{84.6 \tiny($\pm$ 0.62)} & 7.5 \tiny($\pm$ 0.44) & 83.7 & 5.4 \\
RanPAC & 89.5 \tiny($\pm$ 1.12) & 4.2 \tiny($\pm$ 0.77) & 81.7 \tiny($\pm$ 1.34) & 8.0 \tiny($\pm$ 0.94) & 77.2 \tiny($\pm$ 0.48) & 4.2 \tiny($\pm$ 0.17) & \underline{75.3 \tiny($\pm$ 0.88)} & 7.2 \tiny($\pm$ 0.26) & 87.4 \tiny($\pm$ 0.77) & 4.4 \tiny($\pm$ 0.34) & 83.1 \tiny($\pm$ 0.32) & \underline{4.7 \tiny($\pm$ 0.13)} & 82.0 & 6.1 \\ 
Fitness & 84.5 \tiny($\pm$ 2.61) & 5.6 \tiny($\pm$ 2.07) & 82.0 \tiny($\pm$ 1.97) & 9.4 \tiny($\pm$ 1.49) & 72.5 \tiny($\pm$ 0.77) & 5.8 \tiny($\pm$ 0.14) & 70.3 \tiny($\pm$ 0.83) & 8.4 \tiny($\pm$ 0.20) & 81.2 \tiny($\pm$ 0.31) & 6.1 \tiny($\pm$ 0.25) & 79.7 \tiny($\pm$ 0.33) & 9.0 \tiny($\pm$ 0.28) & 78.3 & 7.3 \\ 
KEM & 86.2 \tiny($\pm$ 1.91) & 3.8 \tiny($\pm$ 1.16) & 81.0 \tiny($\pm$ 0.98) & \underline{5.6 \tiny($\pm$ 0.87)} & 76.5 \tiny($\pm$ 0.54) & 3.7 \tiny($\pm$ 0.39) & 74.7 \tiny($\pm$ 0.79) & 4.6 \tiny($\pm$ 0.47) & 86.1 \tiny($\pm$ 0.59) & 4.3 \tiny($\pm$ 0.35) & 83.7 \tiny($\pm$ 0.67) & 5.9 \tiny($\pm$ 0.33) & 81.4 & 4.7 \\ 
\midrule
\textbf{FoRo} & 88.3 \tiny($\pm$ 1.18) & \textbf{2.4 \tiny($\pm$ 0.46)} & \textbf{87.7 \tiny($\pm$ 0.83)} & \textbf{3.5 \tiny($\pm$ 0.51)} & \underline{78.6 \tiny($\pm$ 0.65)} & \textbf{3.0 \tiny($\pm$ 0.31)} & \textbf{76.7 \tiny($\pm$ 0.64)} & \textbf{3.7 \tiny($\pm$ 0.19)} & \underline{89.4 \tiny($\pm$ 0.52)} & \textbf{3.4 \tiny($\pm$ 0.23)} & \textbf{86.2 \tiny($\pm$ 0.55)} & \textbf{4.3 \tiny($\pm$ 0.25)} & \textbf{84.5} & \textbf{3.4} \\ 
\bottomrule
\end{tabular}
}
\end{table*}

\subsection{Comparison Results}
Table~\ref{tab:Performance_Comparision} presents a comprehensive comparison of CL methods across three datasets: CIFAR-100, ImageNet-R, and CUB-200. For each dataset, we report results under two incremental learning settings: $T=5$ tasks (coarse-grained) and $T=25$ tasks (fine-grained). Each setting includes two metrics: the average accuracy ($\overline{\text{A}}$) and average forgetting ($\overline{\text{F}}$), following standard evaluation protocols. \looseness=-1

The final two columns on the right summarize the overall average accuracy ($\overline{\text{A}}$) and average forgetting ($\overline{\text{F}}$) across all datasets and task granularities. 
Fitness represents the use of CMA-ES alone, with Eqn.~\eqref{fitness function} serving as the fitness function.
KEM refers to the exclusive use of knowledge encoding mechanism, as detailed in Section~\ref{know. encoding}.

Based on the results in Table~\ref{tab:Performance_Comparision}, we make the following observations:
1) 
FoRo achieves the highest overall performance across all methods. FoRo achieves the best average accuracy (84.5\%) and the lowest average forgetting (3.4\%). This result demonstrates that FoRo balances task adaptability and knowledge retention effectively, delivering high accuracy while maintaining model stability.

2) 
Compared to traditional CL methods (LwF, DSG, and L2P), FoRo exhibits superior robustness against task interference. For example, on the long task sequence of ImageNet-R ($T=25$), FoRo achieves 76.7\% accuracy, significantly outperforming LwF (58.0\%), DSG (66.3\%), and prompt-based L2P (68.0\%). Regarding forgetting, FoRo maintains a rate of only 3.7\%, compared to L2P's 7.6\%. These results indicate that FoRo generalizes more stably when facing large feature distribution shifts or numerous tasks, and it preserves previously acquired knowledge more effectively.

3) 
FoRo demonstrates better scalability and cross-task stability compared to other prompt-based methods such as LayUP, ADAM, and RanPAC. Although these methods combine pre-trained models with task-specific modules, their performance tends to degrade with longer task sequences due to reliance on early-stage optimization. For instance, in CUB-200 with $T=25$, the forgetting of L2P and RanPAC reaches 7.6\% and 7.2\% respectively, while FoRo maintains a lower rate of 4.3\%. In addition, LayUP shows greater performance fluctuation on ImageNet-R, whereas FoRo consistently preserves high accuracy across datasets, suggesting better robustness.

4) 
Without introducing extra training phases or parameter overhead, FoRo outperforms many parameter-efficient fine-tuning approaches. While ADAM fine-tunes part of the pre-trained model using adapters and LayUP enhances generalization via multi-layer contrastive features, both require additional modules and early-stage training. In contrast, FoRo maintains a purely forward-only process without modifying the pre-trained model. It achieves 86.2\% accuracy on CUB-200 with $T=25$, surpassing ADAM (84.6\%) and LayUP (83.4\%), thus offering a favorable trade-off between performance and efficiency.

5) 
FoRo’s design, which integrates prompt adaptation and knowledge encoding, reveals a clear complementary effect. When using CMA-ES alone to optimize prompts (i.e., Fitness), the model reaches 82.0\% accuracy; using only the knowledge encoding mechanism (i.e., KEM) results in 81.4\%. When combined in FoRo, the accuracy increases to 84.5\%, and forgetting drops from 6.0\%--4.7\% to just 3.4\%. This synergy confirms the effectiveness of jointly adapting to new tasks and explicitly accumulating historical knowledge.

\subsection{Ablation Studies}

\textbf{Effectiveness of Components in FoRo.}
We conduct ablation studies on these components as shown in Table~\ref{tab:foro_ablation}. 
Using Entropy (73.2\%) or Activation Discrepancy (71.4\%) alone significantly outperforms FT (57.3\%). When combined, it achieves a score of 78.4\%, which is comparable to L2P (77.6\%, see Table~\ref{tab:Performance_Comparision}).
Secondly, relying solely on the KEM (introduced in Section~\ref{know. encoding}) significantly improves the accuracy from 57.3\% to 81.4\%, demonstrating its effectiveness in retaining old knowledge.
Lastly, by combining Entropy and Activation Discrepancy as the complete fitness function along with KEM, our FoRo achieves the best performance, with an average accuracy of 84.5\% and an average forgetting of 3.4\%. 

\textbf{Why FoRo Performs Well.}
When using CMA-ES alone to adapt prompts, FoRo achieves a strong performance of 78.4, which is comparable to methods such as L2P (77.6). However, this performance is still lower than that of methods that incorporate additional adaptation mechanisms, including LayUP (83.4). With the integration of the knowledge encoding mechanism, where task-specific features are transformed using nonlinear random projection and classifier weights are updated through recursive least squares, FoRo achieves a notable performance gain. This result highlights the importance of combining prompt adaptation with structured knowledge accumulation to support more effective continual learning.

\begin{table}[t]
\newcommand{\tabincell}[2]{\begin{tabular}{@{}#1@{}}#2\end{tabular}}
 \begin{center}
 \caption{Ablations of components in our FoRo.}\label{tab:foro_ablation}
 \begin{threeparttable}
 \LARGE
    \resizebox{0.8\linewidth}{!}{
  \begin{tabular}{ccc|cc} \\
  \multicolumn{3}{c}{} & \multicolumn{2}{c}{Average} \\
 	 \textit{Entropy} & \textit{Discrepancy} & KEM & $\overline{\bm{\text{A}}}$ (\%, $\uparrow$) & $\overline{\bm{\text{F}}}$ (\%, $\downarrow$) \\
    \cmidrule{1-5}       
        ~ & FT & ~ & 57.3  & 20.2  \\ 
        \checkmark & ~ & ~ & 73.2  & 7.6  \\ 
        ~ & \checkmark & ~ & 71.4  & 8.4  \\ 
        ~ & ~ & \checkmark & 81.4  & 4.7  \\ 
        \checkmark & \checkmark & ~ & 78.4  & 7.3  \\ 
        \checkmark & ~ & \checkmark & 83.7  & 3.9  \\ 
        ~ & \checkmark & \checkmark & 83.2  & 4.3  \\ 
    \cmidrule{1-5}
        \checkmark & \checkmark & \checkmark & \textbf{84.5}  & \textbf{3.4} \\ 
	\end{tabular}
	}
	 \end{threeparttable}
	 \end{center}
  \vspace{-0.1in}
\end{table}   

\begin{figure*}[t]
    \centering
    \includegraphics[width=0.65\linewidth]{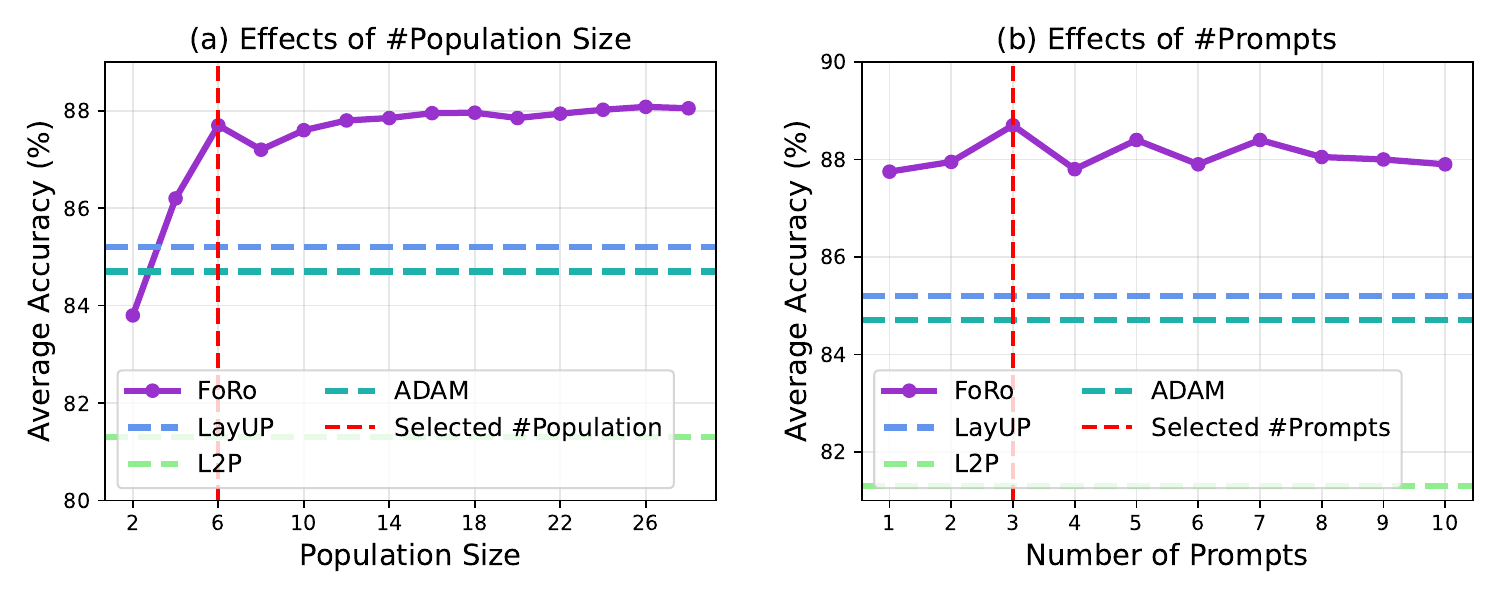}
    \caption{Parameter sensitivity analyses of our FoRo. Experiments are conducted on CIFAR-100 with $T=25$.
    }
    \label{fig:ablation}
\end{figure*}

\textbf{Effects of Population Size in CMA-ES.}
We evaluate the performance of FoRo with different population sizes \( K \) ranging from 2 to 28. As shown in Fig.~\ref{fig:ablation}(a), the performance of FoRo tends to stabilize when \( K > 10 \). Notably, when \( K = 4 \), FoRo outperforms LayUP and ADAM, with an accuracy of 86.2\% compared to 85.2\% for LayUP. 
These results demonstrate the effectiveness of FoRo with smaller population sizes (smaller \( K \) values lead to higher efficiency).

\textbf{Effects of Number of Prompts.}
In FoRo, we add prompt embeddings to the input layer. Here, we evaluate the performance of FoRo with different numbers of prompts, ranging from 1 to 10. As shown in Fig.~\ref{fig:ablation}(b), the performance of FoRo varies only slightly with different numbers of prompts, indicating that it is not sensitive to the number of prompts. Therefore, for all major experiments, we simply fix the number of prompts to 3 without fine-tuning, as it is usually impractical to obtain data for parameter tuning in real-world scenarios.

\subsection{Computational Complexity Analyses.}
We study the effects of different $K$ values (population size in CMA-ES) on the performance and efficiency of FoRo. Table~\ref{tab:computation_complexity} details the $\overline{\bm{\text{A}}}_{25}$, $\overline{\bm{\text{F}}}$, run time and memory usage for various methods on CIFAR-100. The `-' symbol indicates that the number of BP is determined by the number of epochs during the base training phase, such as RanPAC, which is set to 20 epochs.

FoRo consistently maintains high average accuracy and low forgetting rate across different values of \(K\). For instance, at $K=2$, FoRo surpasses RanPAC in accuracy. With \(K = 6\), FoRo achieves an average accuracy of 87.7\%, significantly higher than ADAM (84.7\%) and LayUP (84.9\%), and also surpasses L2P (81.3\%). Additionally, FoRo exhibits a low average forgetting rate of around 3.5\%, much lower than ADAM's 7.1\% and LayUP's 6.9\%. This demonstrates that FoRo effectively maintains high accuracy and significantly reduces knowledge forgetting across long task sequences. \looseness=-1

In terms of resource consumption, FoRo shows significant advantages. 
Although FoRo requires more forward passes compared to methods like LayUP and L2P, it eliminates the need for backpropagation, significantly reducing memory usage.
For example, with \(K = 4\), FoRo's memory usage is only 1,490 MB, far lower than ADAM's 2,257 MB and RanPAC's 2,812 MB. 
Furthermore, FoRo completes tasks in 915 seconds, faster than ADAM (1,502 seconds) and RanPAC (1,585 seconds). 
Compared to L2P (1,691 seconds), FoRo takes slightly less time while achieving superior performance with higher accuracy (87.7\% vs 81.3\%) and lower forgetting rate (3.5\% vs 10.2\%), demonstrating an excellent balance between efficiency and effectiveness.

\begin{table}[t]
\newcommand{\tabincell}[2]{\begin{tabular}{@{}#1@{}}#2\end{tabular}}
 \begin{center}
 \caption{Comparisons \wrt computation complexity. FP/BP refers to forward/backward propagation. \#FP and \#BP are numbers counted for processing a single sample. $\overline{\bm{\text{A}}}_{25}$ and $\overline{\bm{\text{F}}}$ are average results on CIFAR-100 with $T=25$. The Wall-Clock Time (seconds) and Memory Usage (MB) are measured on a single RTX 4090 GPU. The population size $K$ in CMA-ES works well across the range $K\in[2,6]$.}
    \label{tab:computation_complexity}
 \begin{threeparttable}
 \LARGE
    \resizebox{1.0\linewidth}{!}{
  \begin{tabular}{l|ccc|cc|cc}
  Method & BP & \#FP & \#BP & $\overline{\bm{\text{A}}}_{25}$ & $\overline{\bm{\text{F}}}$  & Run Time & Memory\\
  \cmidrule{1-8}
      FT & \checkmark &  1 & 1 & 55.7 & 32.7 & 1,807 & 4,058  \\
      LwF & \checkmark & 1 & 1 & 73.6 & 10.7 & 2,158 & 6,720 \\
      L2P & \checkmark &  1 & 1 & 81.3 & 10.2 & 1,691 & 3,165\\ \midrule
      LayUP & $\Delta$ &  1 & - & 84.9 & 6.9 & 1,735 & 3,106\\
      ADAM & $\Delta$ &  1 & - & 84.7 & 7.1 & 1,502 & 2,257\\
      RanPAC & $\Delta$ &  1 & - & 81.7 & 8.0 & 1,585 & 2,812\\ \midrule
      KEM & $\times$ &  1 & 0 & 81.4 & 4.7 & 241 & 1,457\\
      FoRo ($K\small{=}2$) & $\times$ & 2 & 0 & 83.8 & 4.5 & 457 & 1,478\\
      FoRo ($K\small{=}4$) & $\times$ & 4 & 0 & 86.2 & 3.9 & 915 & 1,490\\
      FoRo ($K\small{=}6$) & $\times$ & 6 & 0 & 87.7 & 3.5 & 1,446 & 1,541\\
	\end{tabular}
	}
	 \end{threeparttable}
	 \end{center}
  \vspace{-0.1in}
\end{table}  

\subsection{Further Analysis and Discussion}
\subsubsection{Long Task Sequences.}
\begin{figure}[t]
    \centering
    \includegraphics[width=1\linewidth]{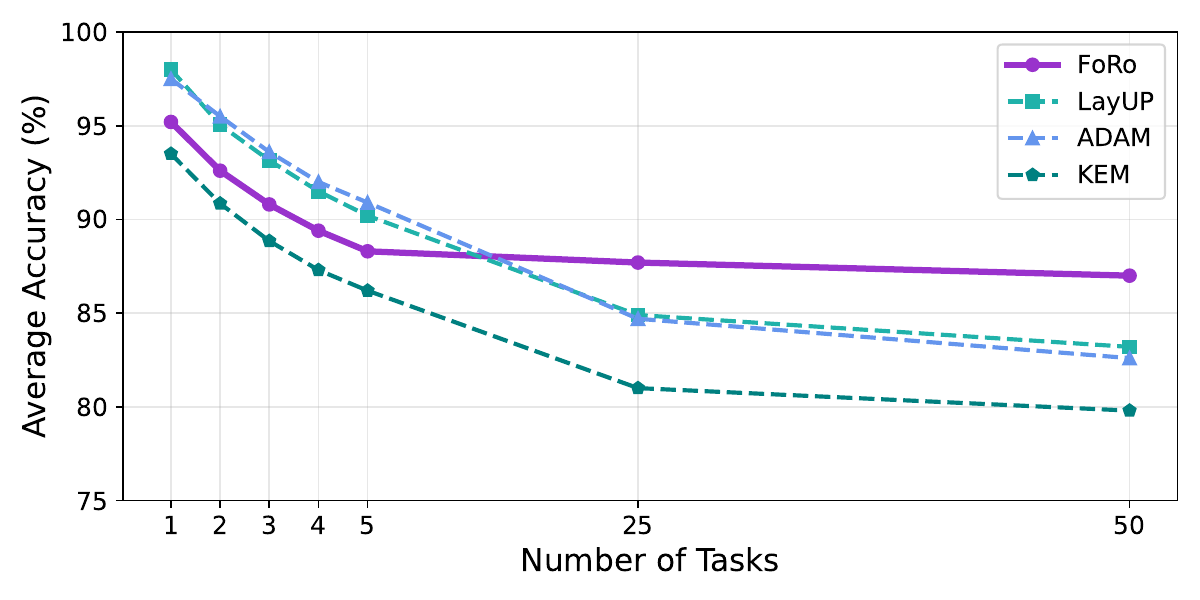}
    \caption{Comparison of different task counts $T$.
    }
    \label{fig:average_acc_tasks}
\end{figure}

We conduct experiments on CIFAR-100 with long task sequences, setting the number of tasks to [5, 25, 50], as shown in Fig.~\ref{fig:average_acc_tasks}. As the number of tasks increases, FoRo maintains high average accuracy and low average forgetting even with $T=50$ tasks. 
In contrast, LayUP and ADAM show higher average forgetting. While they encode knowledge implicitly via class prototypes, this approach struggles in long task sequences to mitigate increasing task interference.
As the number of tasks increases further, their performance is expected to degrade even more.
KEM struggles to maintain performance due to its inadequate representation capabilities, which are insufficient to retain task-specific knowledge effectively.

\subsubsection{Effects of trade-off parameter $\lambda$ in Eqn.~\ref{fitness function}.} In our experiments, we conduct a thorough investigation into the sensitivity of the trade-off parameter \(\lambda\). We experiment with \(\lambda\) values ranging from \{0.1, 0.3, 0.5, 0.7, 0.9\}. As shown in Table~\ref{tab:lambda ablation}, FoRo maintained comparable average accuracy across different \(\lambda\) values, indicating that it is largely insensitive to changes in \(\lambda\). However, the best performance was observed when \(\lambda\) was set to 0.3.

\begin{table}[t]
\newcommand{\tabincell}[2]{\begin{tabular}{@{}#1@{}}#2\end{tabular}}
 \begin{center}
 \caption{Sensitivity analyses of $\lambda$. We report results on ImageNet-R with $T=5$.}
 \label{tab:lambda ablation}
 \begin{threeparttable}
 \resizebox{1.0\linewidth}{!}{
  \begin{tabular}{lccccc}
  ~ & $\lambda=0.1$ & $\lambda=0.3$ & $\lambda=0.5$ & $\lambda=0.7$ & $\lambda=0.9$ \\
  \cmidrule{1-6}
      $\overline{\bm{\text{A}}}$ (\%, $\uparrow$) & 77.8 & \textbf{78.6} & 78.3 & 78.5 & 78.2 \\
      $\overline{\bm{\text{F}}}$ (\%, $\downarrow$) & 3.4 & 3.0 & 3.2 & 3.1 & 3.3 \\
	\end{tabular}
 }
	 \end{threeparttable}
	 \end{center}
\end{table}

\subsubsection{Scaling with NRP Size}\label{sec:nrp}
Table~\ref{tab:nrp} evaluates the influence of NRP size on the average accuracy $\overline{\bm{\text{A}}}_5$ for CIFAR-100 with \( T=5 \) tasks. The results indicate a steady improvement in accuracy as the NRP size increases, achieving the highest performance (88.2\%) at an NRP size of 8000. However, further enlarging the NRP size to 10,000 or 15,000 introduces marginal performance degradation, potentially due to redundancy or noise in the projected feature space.
Based on this analysis, we adopt an NRP size of 8192 throughout our experiments. This choice is close to the observed performance peak while offering better computational efficiency and numerical stability.


\begin{table}[ht]
\centering
\caption{The impact of NRP size.}
\label{tab:nrp}
\resizebox{1.0\linewidth}{!}{
\begin{tabular}{c|c|c|c|c|c|c}
\toprule
NRP Size & 1,000 & 2,000 & 5,000 & 8,000 & 10,000 & 15,000 \\
\midrule
$\overline{\bm{\text{A}}}_5$ (\%, $\uparrow$) & 86.1 & 87.0 & 87.8 & \textbf{88.2} & 87.9 & 87.5 \\
\bottomrule
\end{tabular}
}
\end{table}

\balance
\section{Conclusion}

We present FoRo, a forward-only, gradient-free continual learning method that operates without modifying the pre-trained model, making it well-suited for resource-constrained settings. FoRo combines evolutionary prompt tuning and a recursive knowledge encoding mechanism to enable efficient task adaptation and classifier updates without backpropagation. Experiments demonstrate superior performance in accuracy, forgetting reduction, and efficiency. 
Owing to its gradient-free optimization and data-free updates, FoRo offers a promising solution for continual learning in multimedia scenarios that demand both efficiency and adaptability.

\section{Acknowledgments}
The work of Jianhua Tang was supported in part by the National Key R\&D Program of China under Grant 2024YFE0200500, in part by the Guangdong Basic and Applied Basic Research Foundation under Grant 2024A1515012615, in part by the Department of Science and Technology of Guangdong Province under Grant 2021QN02X491, and in part by the Foundation and Application Research Grant of Guangzhou under Grant 2025A04J2742.

\bibliographystyle{ACM-Reference-Format}
\bibliography{sample-base}


\end{document}